\documentclass{article}

\PassOptionsToPackage{numbers, compress}{natbib}

\usepackage[preprint]{neurips_2025}

\usepackage[utf8]{inputenc} 
\usepackage[T1]{fontenc}    
\usepackage{hyperref}       
\usepackage{url}            
\usepackage{booktabs}       
\usepackage{amsfonts}       
\usepackage{graphicx}       
\usepackage{nicefrac}       
\usepackage{microtype}      
\usepackage{xcolor}         
\usepackage{amsmath}
\usepackage{multirow}
\usepackage{caption}  
\usepackage{array}
\usepackage{algorithm}
\usepackage{algpseudocode}
\usepackage{amsmath}
\usepackage[utf8]{inputenc}
\usepackage{pifont}
\usepackage{bm}
\usepackage{xcolor}
\definecolor{myteal}{HTML}{15797A}
\usepackage{wrapfig}

\title{DMax: Aggressive Parallel Decoding for dLLMs}

\author{%
  Zigeng Chen,  \ Gongfan Fang, \ Xinyin Ma, \ Ruonan Yu,
\ Xinchao Wang\thanks{Correspoding Author} \\
  National University of Singapore\\
  \texttt{zigeng99@u.nus.edu, xinchao@nus.edu.sg} \\
}

\begin{document}

\maketitle

\begin{abstract}

  We present DMax, a new paradigm for efficient diffusion language models (dLLMs). It mitigates error accumulation in parallel decoding, enabling aggressive decoding parallelism while preserving generation quality. Unlike conventional masked dLLMs that decode through a binary mask-to-token transition, DMax reformulates decoding as a progressive self-refinement from mask embeddings to token embeddings. At the core of our approach is On-Policy Uniform Training, a novel training strategy that efficiently unifies masked and uniform dLLMs, equipping the model to recover clean tokens from both masked inputs and its own erroneous predictions. Building on this foundation, we further propose Soft Parallel Decoding. We represent each intermediate decoding state as an interpolation between the predicted token embedding and the mask embedding, enabling iterative self-revising in embedding space. Extensive experiments across a variety of benchmarks demonstrate the effectiveness of DMax. Compared with the original LLaDA-2.0-mini, our method improves TPF on GSM8K from 2.04 to 5.47 while preserving accuracy. On MBPP, it increases TPF from 2.71 to 5.86 while maintaining comparable performance. On two H200 GPUs, our model achieves an average of 1,338 TPS at batch size 1. Code is available at: \textcolor{magenta}{\url{https://github.com/czg1225/DMax}}
  
\end{abstract}

\begin{figure*}[h!]
\centering
\makebox[\textwidth][c]{\includegraphics[width=1.0\textwidth]{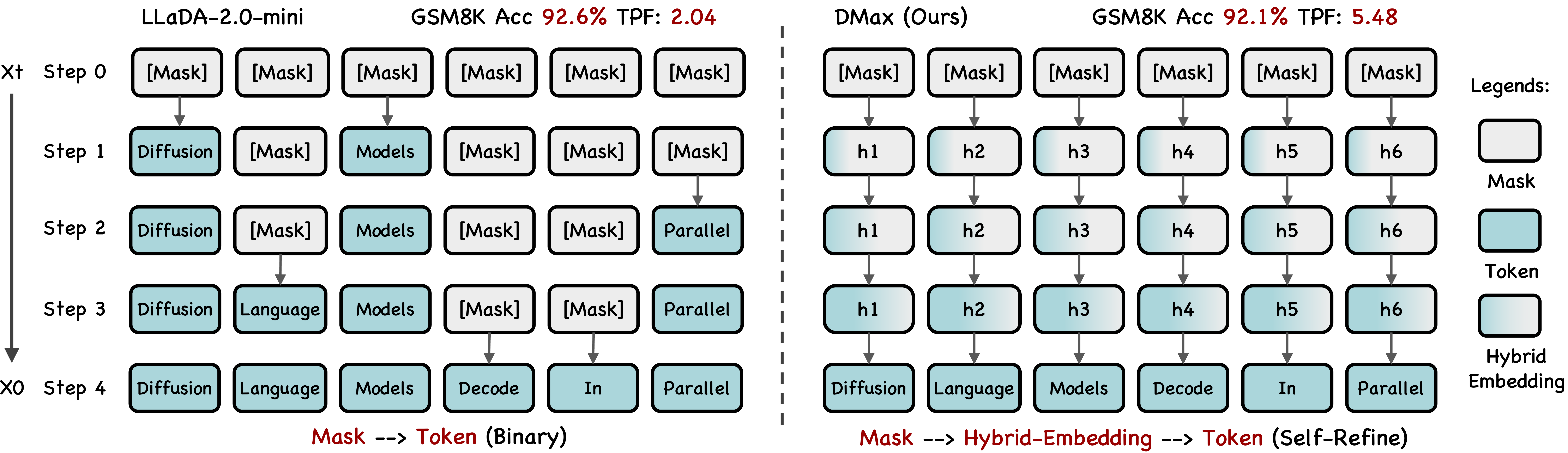}}
\caption{Comparison between the original LLaDA-2.0-mini and our proposed DMax. Unlike the original binary mask-to-token decoding process, DMax introduces a self-revising mask-to-hybrid-embedding-to-token process, enabling highly parallel decoding without accuracy dropping.}
\label{fig_intro}
\end{figure*}

\section{Introduction}

Recently, Diffusion Language Models (dLLMs) \cite{yi2024diffusion,yu2025discrete,zhang2025survey,li2025survey,ni2025diffusion,zhou2026dllm} have emerged as a compelling alternative to the long-standing dominance of Autoregressive Language Models (AR-LLM) \cite{achiam2023gpt, bai2023qwen,grattafiori2024llama} in text generation. The primary allure of dLLMs lies in their capacity for parallel decoding, which holds great promise for improving inference efficiency

Despite this promise, the practical decoding parallelism of existing dLLMs \cite{nie2025large, zhu2025llada,ye2025dream,cheng2025sdar,bie2025llada2,yang2025mmada} remains limited, as their performance drops sharply under aggressive parallel decoding. Some prior work has attempted to improve this trade-off through improved decoding \cite{israel2025accelerating, wei2025accelerating, li2025beyond, gwak2025reward, ben2025accelerated,hong2025wide,hu2026residual,xu2025lopa} or distillation strategies \cite{chen2025dparallel,qian2026d3llm,zhang2026t3d,kim2025cdlm}. Nevertheless, these methods do not address the fundamental bottleneck underlying parallel decoding in current dLLM paradigm: error accumulation.

In current mask-based dLLMs, decoding is a binary, one-way mask-to-token process. Once a masked position is decoded into a token, that token is fixed and propagated as context to subsequent decoding steps, with no opportunity for revision. Under highly parallel decoding, erroneous predictions are inevitable. Once such errors are committed, they contaminate future predictions and trigger cascading error accumulation, ultimately leading to semantic collapse. Unlike speculative decoding \cite{leviathan2023fast,cai2024medusa,li2024eagle}, dLLMs lack a mechanism to recover from incorrect predictions, which fundamentally restricts their performance under highly parallel decoding. Addressing this challenge requires a new dLLM paradigm with an intrinsic capability to revise its own predictions during decoding.

Building on this insight, we propose DMax, a novel paradigm that reformulates the binary mask-to-token decoding process into a self-revising transformation in the embedding space. Central to our approach is On-Policy Uniform Training (OPUT), a training recipe that efficiently extends a pretrained masked diffusion language model into a self-corrective uniform diffusion language model while preserving its original mask denoising capability. Unlike conventional uniform diffusion training that constructs noisy sequences by randomly sampling tokens from the vocabulary, OPUT samples noisy inputs on-policy from the model's own predictive distribution.  This substantially bridges the train-inference gap and enables the model to effectively learn to correct its own potential prediction errors. Building upon OPUT, we further present Soft Parallel Decoding (SPD) for inference. Instead of treating decoded tokens as discrete and irrevocable commitments, SPD represents each intermediate decoding state as a hybrid soft embedding, formed by interpolating between the predicted token embedding and the mask embedding according to the model's prediction confidence. This simple design provides the model with confidence priors from previous steps, enabling more robust self-correction.

Using LLaDA-2.0-mini \cite{bie2025llada2}, a state-of-the-art open-source dLLM, as the base model, we validate the effectiveness of our method across multiple widely used benchmarks. On the mathematical reasoning benchmark GSM8K \cite{cobbe2021training}, our method increases tokens per forward (TPF) from 2.04 to 5.48 with only minimal accuracy degradation relative to the original model. On the code generation benchmark MBPP \cite{austin2021program}, it improves TPF from 2.71 to 5.86 while maintaining comparable performance.

In summary, we propose DMax, a novel paradigm that enables highly parallel decoding for dLLMs while preserving strong performance. Our central idea is to mitigate the error accumulation issue caused by the conventional one-way mask-to-token decoding. To realize this, we introduce two key designs: on-policy uniform training and soft parallel decoding. Extensive experiments demonstrate the effectiveness and superiority of our approach. This work establishes a new strong baseline for future research on parallel decoding in dLLMs.

\section{Preliminaries}
We begin by briefly reviewing the diffusion language modeling paradigms, and then highlight the central challenge for highly parallel decoding and introduce our key motivation.

\noindent \textbf{Masked Diffusion Language Models (MDLMs).}
MDLMs \cite{shi2024simplified, austin2021structured,sahoo2024simple,zheng2024masked, lou2023discrete} formulate text generation as a discrete denoising process over token sequences, where clean tokens are progressively replaced by a special \texttt{[MASK]} symbol during corruption. Let $x_0 = (x_0^1, \dots, x_0^L) \in \mathcal{V}^L$ denote a clean sequence of length $L$, where $\mathcal{V}$ is the vocabulary. Given a corrupted sequence $x_t$ at noise level $t \in [0,1]$, the denoising model is trained to recover the original tokens only at masked positions. The standard MDLM objective is
\begin{equation}
\mathcal{L}_{\mathrm{MDLM}}(\theta)
=
-
\mathbb{E}_{x_0,\, t,\, x_t}
\left[
\frac{1}{t}
\sum_{i=1}^{L}
\mathbf{1}(x_t^i = \texttt{[MASK]})
\log p_\theta(x_0^i \mid x_t)
\right].
\label{eq:mdlm-loss}
\end{equation}
At inference time, MDLMs start from a fully masked sequence and iteratively decode masked positions in parallel, with an optional remasking step to enable further refinement.

\vspace{0.5em}
\noindent \textbf{Uniform Diffusion Language Models (UDLMs).}
UDLMs \cite{schiff2025simple,sahoo2025diffusion,sekhar2026scaling} generalize the corruption process by replacing tokens with uniformly sampled vocabulary tokens rather than a dedicated \texttt{[MASK]} symbol. As a result, the model is trained to recover clean tokens from arbitrary noisy token inputs, instead of only from masked positions. A standard UDLM training objective is
\begin{equation}
\mathcal{L}_{\mathrm{UDLM}}(\theta)
=
-
\mathbb{E}_{x_0,\, t,\, x_t}
\left[
\sum_{i=1}^{L}
\log p_\theta(x_0^i \mid x_t)
\right].
\label{eq:udlm-loss}
\end{equation}
During inference, UDLMs typically start from a fully noisy sequence sampled uniformly from the vocabulary and iteratively update all positions.

\noindent \textbf{Error accumulation in MDLMs.}
Existing dLLMs based on the MDLM paradigm degrade sharply under highly parallel decoding, which limits the practical speedup. The main reason behind it is error accumulation. MDLM decoding follows a binary mask-to-token process: each position is either a mask token or a committed token. Once a masked position is decoded, its prediction is treated as fixed context for subsequent steps. Early mistakes cannot be revised, and instead propagate through later denoising steps as erroneous context.

\noindent \textbf{UDLMs as a Promising Solution.}
In contrast, UDLMs are trained to denoise from arbitrary vocabulary tokens rather than only from \texttt{[MASK]}, so all positions can be re-evaluated at every decoding step. This token-to-token denoising mechanism naturally enables self-correction and improves robustness to prediction errors. However, UDLM decoding typically starts from a fully random sequence, which makes denoising harder and leads to very unstable generation.

\noindent \textbf{Unify the Strengths of MDLMs and UDLMs.}
Motivated by this trade-off, we propose to unify the strengths of both paradigms. Specifically, we retain a fully masked sequence as the initialization of UDLM decoding to preserve stability, while continuing to re-predict all tokens that have been decoded from \texttt{[MASK]} at every subsequent step. This design combines the stable initialization with the self-revising capability, enabling a more robust parallel decoding process.

\section{Methodology}

\begin{figure*}[t!]
\centering
\includegraphics[width=\textwidth]{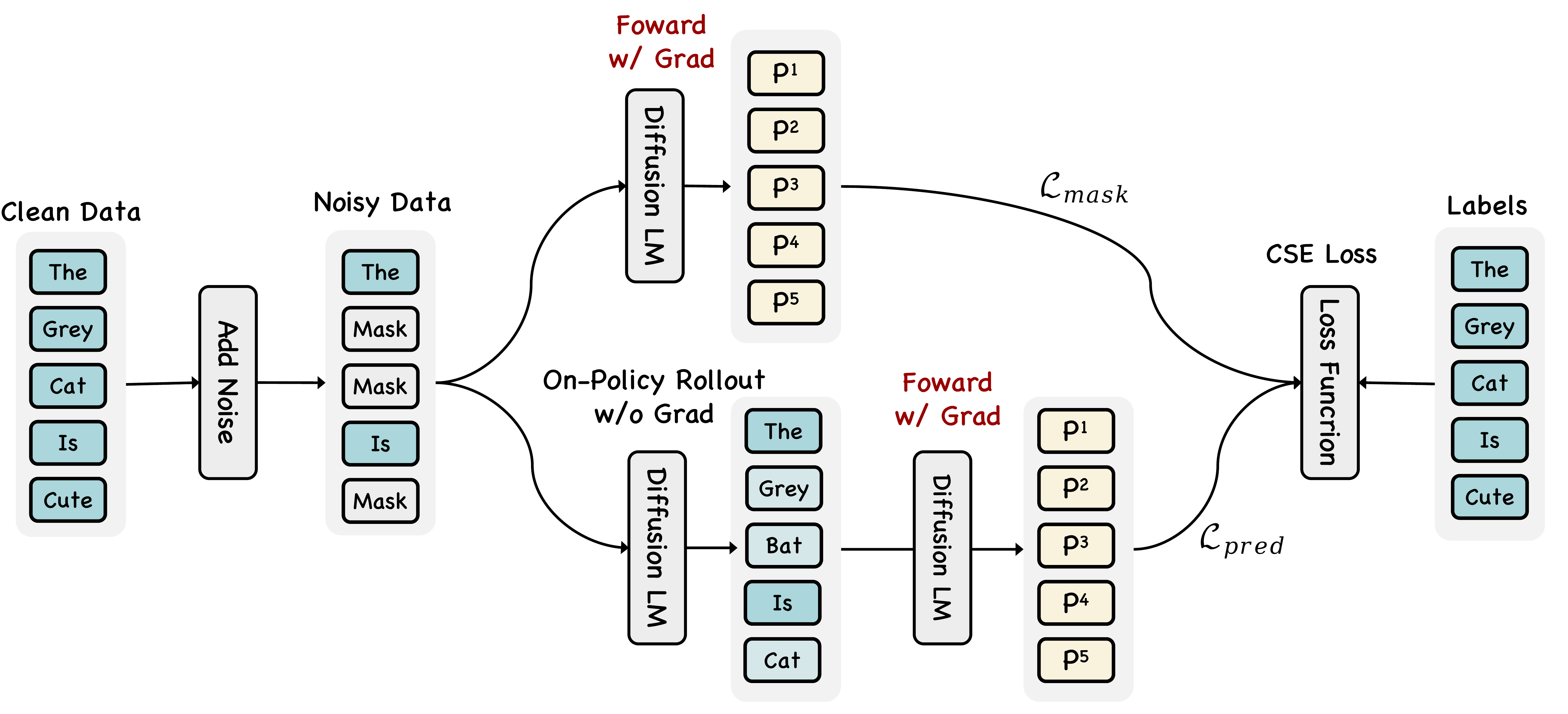}
\caption{Overview of the proposed On-Policy Uniform Training.}
\label{fig_train}
\end{figure*}

\subsection{On-Policy Uniform Training}

A practical way to achieve this goal is to extend a pretrained MDLM into a UDLM. Accordingly, our first objective is to endow a pretrained MDLM with the self-revision capability of UDLMs while preserving its original mask denoising ability.

\noindent \textbf{Extending MDLM toward UDLM is Nontrivial.} This is nontrivial because the training objective of UDLMs differs substantially from that of MDLMs and is considerably harder to optimize. In the standard UDLM training paradigm, a clean sequence is first corrupted by randomly selecting a subset of positions and replacing the selected tokens with tokens sampled uniformly from the vocabulary. The resulting noisy sequence is then used as model input, and the model is trained to recover the original clean sequence.

However, this training strategy is often unstable in practice and tends to yield suboptimal performance. A key reason is that uniformly sampled tokens lie far outside the natural language manifold, producing highly unnatural corrupted inputs. As a result, the model must spend substantial capacity merely learning to map these corrupted sequences back toward plausible language, rather than directly acquiring effective language modeling and self-correction behaviors. More importantly, this corruption process introduces a severe train--inference mismatch. Unlike conventional UDLMs, our paradigm first predicts tokens from masked positions in parallel and then iteratively refines its own predictions. Consequently, the noisy sequences encountered at inference time are sampled from the model's own output distribution rather than from a uniform vocabulary distribution. This mismatch hinders self-correction and leads to ineffective training.

\noindent \textbf{On-Policy Uniform Training.} To address these issues, we propose On-Policy Uniform Training (OPUT), a simple yet effective method for equipping MDLMs with self-corrective denoising capability. The core idea is to construct training inputs using noisy sequences sampled on-policy from the model's own predictive distribution, rather than from a uniform vocabulary distribution, thereby bridging the train--inference gap. The overview of the training procedure is shown in Figure~\ref{fig_train}.

\noindent \textbf{Training Procedure.}
Let $M_\theta$ denote a pretrained diffusion language model built on the MDLM paradigm, parameterized by $\theta$. 
We further adapt $M_\theta$ on a training dataset $\mathcal{D}$ of clean sequences $x_0 = (x_0^1, \dots, x_0^L)$. At each training iteration, we first sample a corruption level $t \sim \mathrm{Uniform}(t_l, t_h)$, where $t_l$ and $t_h$ denote the lower and upper bounds of the noise level, respectively. Given a clean sequence $x_0 \sim \mathcal{D}$, we construct a masked noisy sequence $x_t^{(m)}$ by independently replacing each token with \texttt{[MASK]} with probability $t$.

We feed $x_t^{(m)}$ into $M_\theta$ and predict all masked positions in parallel. 
By sampling from the model's predictive distribution at masked positions, we obtain a predicted noisy sequence $x_t^{(p)}$, defined as
\begin{equation}
x_t^{(p),i}
=
\begin{cases}
x_t^{(m),i}, & \text{if } x_t^{(m),i} \neq \texttt{[MASK]}, \\[4pt]
\hat{x}^i,\quad \hat{x}^i \sim p_\theta(\cdot \mid x_t^{(m)}), & \text{if } x_t^{(m),i} = \texttt{[MASK]}.
\end{cases}
\label{eq:predicted_noisy_sequence}
\end{equation}
Importantly, $x_t^{(p)}$ is sampled using the current model parameters at each iteration, making this a strictly on-policy rollout process.

Next, we perform two forward passes, using the masked noisy sequence $x_t^{(m)}$ and the predicted noisy sequence $x_t^{(p)}$ as inputs, respectively:
\begin{equation}
p_\theta^{(m)}(\cdot \mid x_t^{(m)}) = M_\theta(x_t^{(m)}), 
\qquad
p_\theta^{(p)}(\cdot \mid x_t^{(p)}) = M_\theta(x_t^{(p)}).
\label{eq:two_forward_passes}
\end{equation}
We then supervise both outputs against the original clean sequence $x_0$ using cross-entropy loss over \emph{all} token positions, regardless of whether a position is masked:
\begin{equation}
\mathcal{L}_{\mathrm{mask}}
=
-
\sum_{i=1}^{L}
\log p_\theta^{(m)}(x_0^i \mid x_t^{(m)}),
\qquad
\mathcal{L}_{\mathrm{pred}}
=
-
\sum_{i=1}^{L}
\log p_\theta^{(p)}(x_0^i \mid x_t^{(p)}).
\label{eq:two_losses}
\end{equation}
The final training objective is
\begin{equation}
\mathcal{L}_{\mathrm{on\mbox{-}policy}}
=
\mathcal{L}_{\mathrm{mask}} + \mathcal{L}_{\mathrm{pred}},
\label{eq:final_loss}
\end{equation}
 By reducing the train--inference mismatch, the proposed OPUT strategy enables a pretrained MDLM to efficiently learn self-correction through limited post-training, while retaining its original mask denoising ability. As a result, the model can correct self-generated errors and effectively mitigate error accumulation under highly parallel decoding. On LLaDA-2.0-mini, our method improves GSM8K accuracy from $78\%$ to $90\%$ under confidence-threshold decoding with a threshold of $0.5$, while also delivering faster decoding.

\begin{figure*}[t!]
\centering
\includegraphics[width=\textwidth]{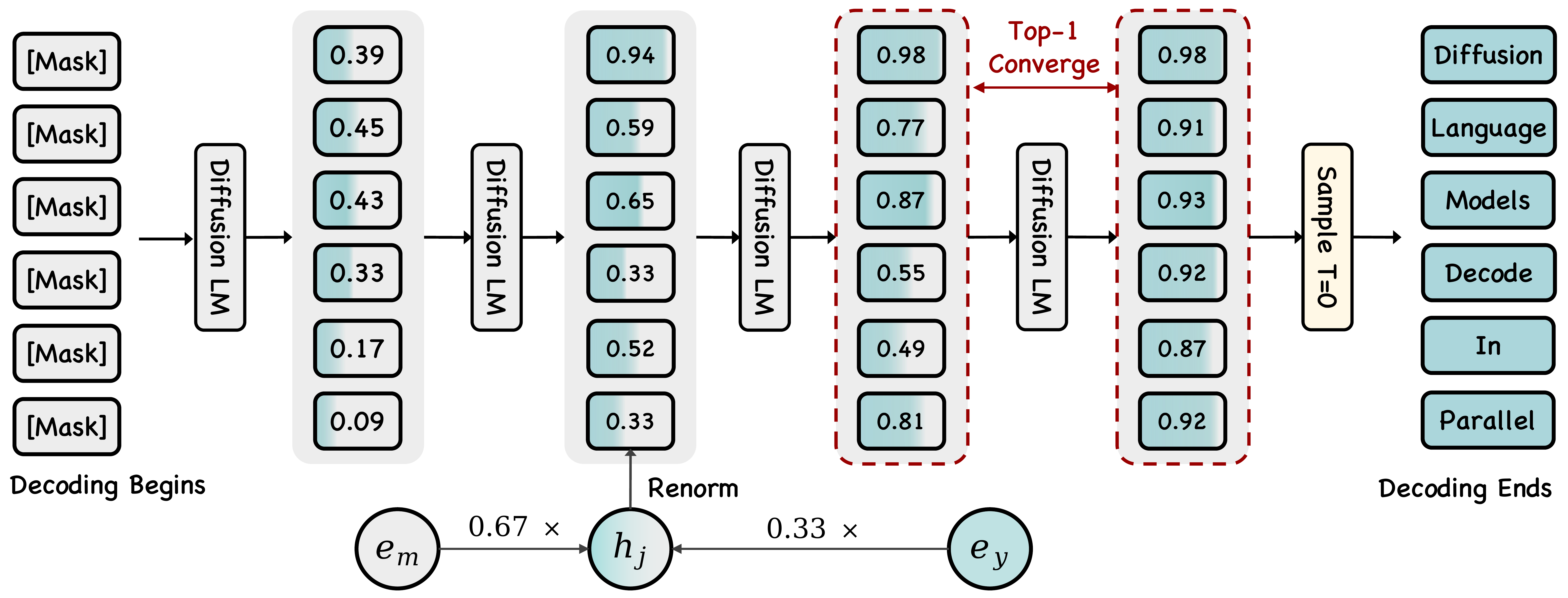}
\caption{Overview of the Soft Parallel Decoding procedure in DMax.}
\label{fig_decode}
\end{figure*}

\subsection{Soft Parallel Decoding}

Although OPUT substantially mitigates error accumulation, it still struggles when many erroneous predictions arise simultaneously within a block. When many positions are decoded in parallel, correlated errors can appear at once, making them difficult to fully correct through iterative refinement. For example, for OPUT-trained LLaDA-2.0-mini, if we decode all masked positions in a block at once using a confidence threshold of \(0\), and then iteratively refine them, the accuracy on GSM8K drops to only \(68\%\).

\noindent \textbf{Soft Parallel Decoding.}
To further enhance self-revising in iterative refinement, we propose soft parallel decoding. The central idea is to preserve predictive uncertainty from earlier iterations and explicitly propagate it to later refinement steps. Concretely, instead of treating intermediate decoding states as discrete tokens, we represent each decoded token as a soft embedding interpolated between the predicted token embedding and the mask embedding. Because the mask embedding naturally encodes maximal uncertainty, this interpolation serves as an explicit carrier of uncertainty across iterations. This enables the model to better distinguish confident predictions from unreliable ones, allowing it to focus on refining low-confidence tokens while avoiding interference from noisy signals.

\noindent \textbf{Decoding Procedure.}
An overview of the decoding process is shown in Figure~\ref{fig_decode}. It follows a block-wise semi-autoregressive process. For each block, we partition its positions into two sets: \emph{mask positions} and \emph{token positions}. At initialization, all positions in the block are mask positions. At each decoding step, we use an aggressive confidence threshold $\tau_{\mathrm{dec}}$ to promote some mask positions into token positions. Specifically, we scan the masked region from left to right and promote only its longest contiguous prefix whose confidence exceeds $\tau_{\mathrm{dec}}$. Once the first mask position with confidence below $\tau_{\mathrm{dec}}$ is encountered, all mask positions to its right remain masked. If no mask position satisfies this criterion, we still promote the leftmost mask position to ensure decoding progress. This design keeps the masked region contiguous and prevents unreliable future tokens on the right from interfering with mask predictions on the left.

At decoding step $t$, every mask position uses the mask embedding as model input:
\begin{equation}
\mathbf{h}^{(t)}_j = \mathbf{e}_{\mathrm{mask}}, \qquad j \in \mathcal{M}^{(t)}.
\label{eq:mask_embedding}
\end{equation}
where $\mathcal{M}^{(t)}$ denotes the set of mask positions at step $t$. 

For each token position $j \in \mathcal{T}^{(t)}$, where $\mathcal{T}^{(t)}$ is the set of token positions, we construct a hybrid embedding from the top-1 prediction at the previous step $t-1$ as the model input. Let
$y^{(t-1)}_j$
denote the top-1 predicted token at position $j$, and let
$\pi^{(t-1)}_j$
be its predicted probability. We assign the remaining probability mass to the mask embedding:
\begin{equation}
\pi^{(t-1)}_{j,\mathrm{mask}} = 1 - \pi^{(t-1)}_j.
\label{eq:mask_probability}
\end{equation}
The unnormalized hybrid embedding is then
\begin{equation}
\tilde{\mathbf{h}}^{(t)}_j
=
\pi^{(t-1)}_j \, \mathbf{e}\!\left(y^{(t-1)}_j\right)
+
\pi^{(t-1)}_{j,\mathrm{mask}} \, \mathbf{e}_{\mathrm{mask}},
\qquad j \in \mathcal{T}^{(t)}.
\label{eq:hybrid_embedding}
\end{equation}

Directly adding high-dimensional embeddings may distort their magnitude and lead to norm collapse. To avoid this issue, we renormalize the hybrid embedding so that its norm matches the probability-weighted sum of the component norms:
\begin{equation}
\mathbf{h}^{(t)}_j
=
\frac{\tilde{\mathbf{h}}^{(t)}_j}{\left\|\tilde{\mathbf{h}}^{(t)}_j\right\|_2}
\left(
\pi^{(t-1)}_j
\left\| \mathbf{e}\!\left(y^{(t-1)}_j\right) \right\|_2
+
\pi^{(t-1)}_{j,\mathrm{mask}} \left\| \mathbf{e}_{\mathrm{mask}} \right\|_2
\right).
\label{eq:renormalized_hybrid_embedding}
\end{equation}
This hybrid embedding serves as a soft intermediate state between decoding steps, explicitly carrying forward the uncertainty of previous predictions.

We regard a block as having converged to a stable state if either of the following conditions holds: (1) the top-1 predictions at all positions remain unchanged for two consecutive decoding steps, or (2) the confidence of every position in the block exceeds a high acceptance threshold $\tau_{\mathrm{acc}}$. Once a block converges, we commit all token positions in the block according to the final predictions and move on to the next block.

\begin{algorithm}[t]
\caption{Soft Parallel Decoding (Block-Wise)}
\label{alg:SPD}
\begin{algorithmic}[1]
\Require Block positions $\mathcal{B}$,\; decoding threshold $\tau_{\mathrm{dec}}$,\; acceptance threshold $\tau_{\mathrm{acc}}$
\State $\mathcal{M}\gets\mathcal{B},\ \mathcal{T}\gets\emptyset,\ \mathbf{h}_j\gets\mathbf{e}_{\mathrm{mask}}\ \forall j\in\mathcal{B}$ \Comment{initialize with fully masked block}
\Repeat
    \State $p_j(\cdot) \gets p_\theta(\cdot \mid \{\mathbf{h}_j\}_{j\in\mathcal{B}}),\ \forall j\in\mathcal{B}$
    \State $\hat{y}_j\gets \arg\max_{y} p_j(y),\ \ c_j\gets p_j(\hat{y}_j),\ \forall j\in\mathcal{B}$
    \State $\mathcal{P}\gets$ longest contiguous prefix in $\mathcal{M}$ such that $c_j>\tau_{\mathrm{dec}}$ for all $j\in\mathcal{P}$
    \If{$\mathcal{P}=\emptyset$}
        \State $\mathcal{P}\gets \{\text{leftmost position in }\mathcal{M}\}$
    \EndIf
    \State $\mathcal{T}\gets\mathcal{T}\cup\mathcal{P},\qquad \mathcal{M}\gets\mathcal{B}\setminus\mathcal{T}$
    \State $\mathbf{h}_j\gets\mathbf{e}_{\mathrm{mask}},\ \forall j\in\mathcal{M}$ \Comment{Eq.~\eqref{eq:mask_embedding}}
    \State $\mathbf{h}_j\gets \mathrm{HybridEmb}(\hat{y}_j,c_j,\mathbf{e}_{\mathrm{mask}}),\ \forall j\in\mathcal{T}$ \Comment{Eqs.~\eqref{eq:mask_probability}--\eqref{eq:renormalized_hybrid_embedding}}
\Until{$\hat{y}^{(t)}_j=\hat{y}^{(t-1)}_j,\ \forall j\in\mathcal{B}$ \textbf{or} $\min_{j\in\mathcal{B}} c_j>\tau_{\mathrm{acc}}$} \Comment{block converges}
\State \Return $\hat{y}_j,\ \forall j\in\mathcal{B}$ \Comment{commit the block}
\end{algorithmic}
\end{algorithm}

By interpolating prediction-mask embeddings as intermediate states, the model receives an explicit uncertainty prior before every forward pass, leading to substantially more robust parallel decoding. On OPUT-trained LLaDA-2.0-mini, under the highly aggressive setting of $\tau_{\mathrm{dec}}=0$, soft parallel decoding improves GSM8K accuracy from $68\%$ to $90\%$ while achieving a higher speedup. 

\noindent \textbf{OPUT as a Prerequisite.} Notably, soft parallel decoding must be used together with OPUT-trained models. OPUT trains the model to recover the correct target not only from masked inputs, but also from its own sampled predictions. As a result, the model learns a consistent mapping from both mask embeddings and self-predicted token embeddings toward the correct output, which makes interpolation between them meaningful. In contrast, applying soft parallel decoding to a standard diffusion language model without OPUT leads to catastrophic performance collapse.

\begin{table*}[t]
    \centering
    \small
    \caption{Comparison with the original model and different baselines. For our DMax-Math model, we set the decoding threshold to 0.5; for the DMax-Coder model, we set it to 0.65. In addition to TPF, TPS, and accuracy, we also report the AUP score to provide a more comprehensive evaluation of parallel decoding performance. All evaluations are under zero-shot and a batch size of 1.}
    \resizebox{0.95\linewidth}{!}{
    \begin{tabular}{>{\raggedright\arraybackslash}p{3.0cm} p{4.5cm} c c c c}
    \toprule
    \multirow{1}{*}{\textbf{Benchmark}} &
    \multirow{1}{*}{\textbf{Method}} &
    \textbf{TPF $\uparrow$} & \textbf{TPS $\uparrow$} &
    \textbf{Acc. $\uparrow$} & \textbf{AUP Score $\uparrow$} \\
    \midrule
    \multicolumn{6}{c}{\textbf{\textit{Math \& Reasoning Benchmarks}}} \\
    \midrule
    \multirow{5}{*}{\textbf{GSM8K}}
    & LLaDA-2.0-mini & 2.04 & 512 & 92.6\% & 340 \\
    & Hierarchical Decoding & 2.44 & 577 & 91.6\% & 357 \\
    & dParallel SFT & 2.79 & 721 & 92.3\% & 395 \\
    & Uniform Diffusion Training & 2.26 & 493 & 68.7\% & 0 \\
    & \textcolor{myteal}{\textbf{DMax-Math}} & \textcolor{myteal}{\textbf{5.48}} & \textcolor{myteal}{\textbf{1258}} & 92.1\% & \textcolor{myteal}{\textbf{557}} \\
    \midrule

    \multirow{5}{*}{\textbf{MATH500}}
    & LLaDA-2.0-mini & 2.58 & 626 & 75.8\% & 257 \\
    & Hierarchical Decoding & 3.01 & 669 & 73.0\% & 268 \\
    & dParallel SFT & 3.42 & 823 & 75.8\% & 310 \\
    & Uniform Diffusion Training & 2.43 & 530 & 33.6\% & 0 \\
    & \textcolor{myteal}{\textbf{DMax-Math}} & \textcolor{myteal}{\textbf{5.94}} & \textcolor{myteal}{\textbf{1286}} & 75.4\% & \textcolor{myteal}{\textbf{507}} \\
    \midrule

    \multirow{5}{*}{\textbf{Minerva-Algebra}}
    & LLaDA-2.0-mini & 3.01 & 755 & 91.4\% & 363 \\
    & Hierarchical Decoding & 3.40 & 787 & 90.6\% & 382 \\
    & dParallel SFT & 3.91 & 943 & 91.4\% & 430 \\
    & Uniform Diffusion Training & 2.55 & 551 & 42.7\% & 0 \\
    & \textcolor{myteal}{\textbf{DMax-Math}} & \textcolor{myteal}{\textbf{7.03}} & \textcolor{myteal}{\textbf{1492}} & 91.5\% & \textcolor{myteal}{\textbf{658}} \\
    \midrule

    \multirow{5}{*}{\textbf{ASDIV}}
    & LLaDA-2.0-mini & 2.03 & 512 & 92.8\% & 354 \\
    & Hierarchical Decoding & 2.43 & 528 & 92.5\% & 366 \\
    & dParallel SFT & 2.72 & 663 & 93.0\% & 459 \\
    & Uniform Diffusion Training & 2.51 & 515 & 80.8\% & 0 \\
    & \textcolor{myteal}{\textbf{DMax-Math}} & \textcolor{myteal}{\textbf{5.62}} & \textcolor{myteal}{\textbf{1172}} & 92.5\% & \textcolor{myteal}{\textbf{556}} \\
    \midrule

    \multicolumn{6}{c}{\textbf{\textit{Code Generation Benchmarks}}} \\
    \midrule

    \multirow{5}{*}{\textbf{HumanEval-Instruct}}
    & LLaDA-2.0-mini & 4.38 & 1044 & 84.2\% & 369 \\
    & Hierarchical Decoding & 4.67 & 1014 & 81.1\% & 379 \\
    & dParallel SFT & 5.12 & 1229 & 76.8\% & 394 \\
    & Uniform Diffusion Training & 2.93 & 628 & 15.2\% & 0 \\
    & \textcolor{myteal}{\textbf{DMax-Coder}} & \textcolor{myteal}{\textbf{7.36}} & \textcolor{myteal}{\textbf{1557}} & 83.5\% & \textcolor{myteal}{\textbf{637}} \\
    \midrule

    \multirow{5}{*}{\textbf{MBPP-Instruct}}
    & LLaDA-2.0-mini & 2.71 & 662 & 80.6\% & 276 \\
    & Hierarchical Decoding & 2.88 & 685 & 76.6\% & 241 \\
    & dParallel SFT & 3.66 & 880 & 74.7\% & 273 \\
    & Uniform Diffusion Training & 2.84 & 608 & 23.4\% & 0 \\
    & \textcolor{myteal}{\textbf{DMax-Coder}} & \textcolor{myteal}{\textbf{5.86}} & \textcolor{myteal}{\textbf{1264}} & 79.2\% & \textcolor{myteal}{\textbf{482}} \\
    \bottomrule
    \end{tabular}
    }
    \label{tab:main}
\end{table*}

\section{Experiments}

\subsection{Experimental Setup}

\noindent \textbf{Implementation Details.}
We build our method on LLaDA-2.0-mini \cite{bie2025llada2}, a state-of-the-art open-source diffusion language model. During training, we use OPUT with a fixed mask ratio of 0.75. We perform full-parameter fine-tuning for 2 epochs with a batch size of 8, an initial learning rate of $2\times10^{-6}$, and a cosine learning rate schedule. Training follows the block-diffusion setting with a block size of 32. To avoid extra memory overhead, the masked noisy sequence and the predicted noisy sequence are optimized in separate iterations within the same epoch, rather than jointly in a single iteration. Under this setup, we train two models: DMax-Math for mathematical reasoning and DMax-Coder for code generation tasks. All training runs are conducted on 8 H200 GPUs. At inference time, we adopt the proposed SPD decoding strategy under semi-autoregressive block diffusion with a block size of 32. The acceptance threshold for determining whether a block has converged to a stable state is set to $\tau_{\mathrm{acc}}=0.9$.

\noindent \textbf{Training Data.}
We construct all training data through self-distillation. Specifically, we take prompts from public datasets and use LLaDA-2.0-mini to generate responses as training targets. For math, prompts are collected from GSM8K trainset \cite{cobbe2021training}, PRM12K \cite{lightman2023let}, a subset of Numina-Math \cite{li2024numinamath}, and a subset of OpenThoughts \cite{guha2025openthoughts}. For code, prompts are drawn from a subset of OpenCodeInstruct \cite{ahmad2025opencodeinstruct}. Responses are generated with a confidence threshold of 0.95, a block size of 32, and a maximum generation length of 2048 tokens. We discard incomplete generations that do not finish within the length budget. This yields 0.7M math samples and 1.0M code samples. Notably, we do not use any external high-quality responses, all supervision is obtained from the model's own generations.

\noindent \textbf{Evaluation Details.}
We evaluate our method on multiple benchmarks. For mathematical reasoning, we use GSM8K \cite{cobbe2021training}, MATH500 \cite{lightman2023let}, Minerva-Algebra \cite{hendrycks2021measuring}, and ASDIV \cite{miao2021diverse}, and prompt the model to produce chain-of-thought \cite{wei2022chain} reasoning. For code generation, we use the instruction versions of HumanEval \cite{chen2021codex} and MBPP \cite{austin2021program}. All evaluations are conducted with the dInFer \cite{ma2025dinfer} framework on 2 H200 GPUs using tensor parallelism. Besides TPF, TPS, and accuracy, we also report AUP Score \cite{qian2026d3llm} to measure parallel decoding performance. The generation length for all benchmarks is 2048.

\noindent \textbf{Baselines.}
We compare our method against four baselines in terms of both decoding efficiency and generation accuracy: 
(1) LLaDA-2.0-mini, the base model, evaluated with its default confidence-threshold-based parallel decoding strategy using a threshold of 0.95; 
(2) Hierarchical Decoding, an advanced inference strategy that improves parallel decoding via a divide-and-conquer procedure \cite{qihierarchy}. The low threshold is set as 0.2; 
(3) dParallel-SFT, for which we use the LLaDA-2.0-mini-CAP model \cite{bie2025llada2}, where the certainty-forcing loss proposed in dParallel \cite{chen2025dparallel} is incorporated into large-scale supervised fine-tuning to improve decoding parallelism; and 
(4) Uniform Diffusion Training, which continues training the base model using the conventional UDLM objective. In addition to masked noisy sequences, this baseline also replaces tokens with random vocabulary samples to construct uniformly corrupted noisy sequences, while keeping all other training settings identical to those of DMax. During inference, it updates all tokens within a block at every step until convergence.

\begin{figure*}[t!]
\centering
\includegraphics[width=\textwidth]{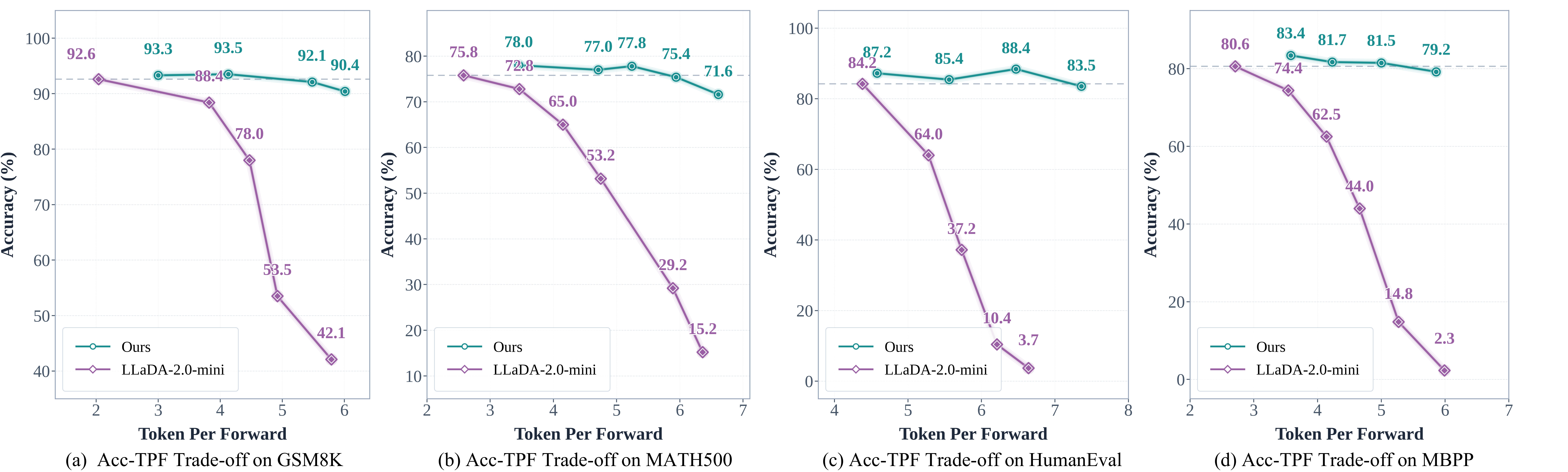}
\caption{Comparison of accuracy-TPF trade-off curves between original LLaDA-2.0-mini model and our method. We present curves on GSM8K, MATH500, HumanEval and MBPP benchmarks.}
\label{fig_tradeoff}
\end{figure*}

\begin{table*}[t!]
    \centering
    \caption{Our proposed new paradigm also improves the model's accuracy at low parallelism.}
    \resizebox{0.75\linewidth}{!}{
    \begin{tabular}{>{\raggedright\arraybackslash}p{2.0cm} p{2.0cm} c c c c}
    \toprule
    \multicolumn{2}{c}{\textbf{Benchmarks}} & 
    \multicolumn{2}{c}{\textbf{LLaDA-2.0-mini}} & 
    \multicolumn{2}{c}{\textbf{DMax}} \\
     \cmidrule(lr){3-4} \cmidrule(lr){5-6}
     &
     &
    \multirow{1}{*}{\textbf{TPF $\uparrow$}} &
    \multirow{1}{*}{\textbf{Acc. $\uparrow$}} &
    \multirow{1}{*}{\textbf{TPF $\uparrow$}} &
    \multirow{1}{*}{\textbf{Acc. $\uparrow$}} \\
    \midrule
    \multicolumn{2}{l}{\textbf{GSM8K}} & 2.04 & 92.6\% & 3.54 \textcolor[HTML]{1A8F91}{(+1.50)} & 93.4\% \textcolor[HTML]{1A8F91}{(+0.8\%)} \\
    \multicolumn{2}{l}{\textbf{MATH500}} & 2.58 & 75.8\% & 3.45 \textcolor[HTML]{1A8F91}{(+0.87)} & 78.0\% \textcolor[HTML]{1A8F91}{(+2.2\%)} \\
    \multicolumn{2}{l}{\textbf{Minerva-Algebra}} & 3.01 & 91.4\% & 4.96 \textcolor[HTML]{1A8F91}{(+1.95)} & 93.6\% \textcolor[HTML]{1A8F91}{(+2.2\%)} \\
    \multicolumn{2}{l}{\textbf{ASDIV}} & 2.03 & 92.8\% & 3.10 \textcolor[HTML]{1A8F91}{(+1.07)} & 93.5\% \textcolor[HTML]{1A8F91}{(+0.7\%)} \\
    \midrule
    \multicolumn{2}{l}{\textbf{HumanEval-Instruct}} & 4.38 & 84.2\% & 4.58 \textcolor[HTML]{1A8F91}{(+0.20)} & 87.2\% \textcolor[HTML]{1A8F91}{(+3.0\%)} \\
    \multicolumn{2}{l}{\textbf{MBPP-Instruct}} & 2.71 & 80.6\% & 3.58 \textcolor[HTML]{1A8F91}{(+0.87)} & 83.4\% \textcolor[HTML]{1A8F91}{(+2.8\%)} \\
    \bottomrule
    \end{tabular}
    }
    \label{tab:low}
\end{table*}

\subsection{Experimental Results}

\noindent \textbf{Aggressive Parallelism While Preserving Accuracy.}
As shown in Table~\ref{tab:main}, compared with the original LLaDA-2.0-mini, our method substantially increases decoding parallelism, improving the average TPF from 2.8 to 6.2 while preserving the original accuracy. In contrast, the other baselines provide only limited gains in parallel decoding. This advantage is further reflected in the AUP Score, where DMax consistently outperforms both the original model and all baselines by a large margin. These results demonstrate that our paradigm enables a much stronger parallel decoding capability than conventional MDLMs. Moreover, on two H200 GPUs, our model achieves a practical inference throughput of over 1000 tokens per second.

\noindent \textbf{On-Policy Training as the Cornerstone.}
Table~\ref{tab:main} also compares our method with conventional uniform diffusion training. The latter neither improves decoding speed nor preserves model quality, instead causing a noticeable performance drop. We find that this failure stems from the large mismatch between the randomly sampled noisy sequences used in training and the model's actual decoding trajectories at inference time. Consequently, the model struggles to revise erroneous predictions while unnecessarily perturbing correct ones, resulting in unstable oscillations within each block. By contrast, our on-policy training samples noisy sequences from the model's own outputs, effectively bridging this train--inference gap and substantially improving self-revision under parallel decoding.

\noindent \textbf{Superior Efficiency--Performance Trade-off.}
Figure~\ref{fig_tradeoff} compares the accuracy--TPF trade-off curves of our method and the original model on GSM8K, MATH500, HumanEval, and MBPP. As TPF increases, the original model suffers a sharp accuracy drop, whereas our method maintains stable performance. For instance, on MATH500, at around 6.5 TPF, our method still retains over 71.6\% accuracy, while the original model falls to 15.2\%. The gap is even larger on code benchmarks: on MBPP, at a similar TPF, our method achieves 79.2\%, whereas the original model drops to only 2.3\%. This superior trade-off stems from the self-revision capability of our paradigm, which effectively mitigates error accumulation under aggressive parallel decoding.

\noindent \textbf{Improved Performance at Low Parallelism.}
By enabling dLLMs to revise their own predictions, our method not only mitigates error accumulation under aggressive parallel decoding, but also improves performance in the low-parallelism regime. Through iterative re-evaluation of earlier predictions, the model can recover from reasoning errors that would otherwise remain on the original decoding path. As shown in Table~\ref{tab:low}, our method consistently improves accuracy by 0.8\%--3.0\% across multiple benchmarks at low parallelism. Importantly, these gains are obtained using only the model's own generated responses as training data, without introducing any external supervision.

\begin{table*}[t!]
    \centering
    \caption{Ablation on different training and inference strategies with different decoding parallelism.}
    \resizebox{0.9\linewidth}{!}{
    \begin{tabular}{c c c c c c c c c}
    \toprule
    \multicolumn{1}{c}{\textbf{Train}} & 
    \multicolumn{2}{c}{\textbf{Inference}} & 
    \multicolumn{2}{c}{\bm{$\tau_{\mathrm{dec}}=0.95$ }} & 
    \multicolumn{2}{c}{\bm{$\tau_{\mathrm{dec}}=0.50$ }} &
    \multicolumn{2}{c}{\bm{$\tau_{\mathrm{dec}}=0.0$ }} \\
    \cmidrule(lr){1-1} \cmidrule(lr){2-3} \cmidrule(lr){4-5} \cmidrule(lr){6-7} \cmidrule(lr){8-9}
    \multirow{2}{*}{\shortstack[c]{\textbf{On-Policy}\\ \textbf{Rollout}}} &
    \multirow{2}{*}{\shortstack[c]{\textbf{Contiguous}\\ \textbf{Prefix}}} &
    \multirow{2}{*}{\shortstack[c]{\textbf{Hybrid}\\ \textbf{Embedding}}} &
    \multirow{2}{*}{\textbf{TPF $\uparrow$}} &
    \multirow{2}{*}{\textbf{Acc. $\uparrow$}} &
    \multirow{2}{*}{\textbf{TPF $\uparrow$}} &
    \multirow{2}{*}{\textbf{Acc. $\uparrow$}} &
    \multirow{2}{*}{\textbf{TPF $\uparrow$}} &
    \multirow{2}{*}{\textbf{Acc. $\uparrow$}} \\
    & & & & & & & & \\
    \midrule
      & &  &2.04  & 92.6\%  &4.47  &78.0\%  & 7.86 & 0.9\% \\
    & \ding{51} & \ding{51} &1.04  &0.0\%   & 1.73 & 0.0\% &  5.39&0.0\%  \\
    \ding{51} &  &  & 2.95 & 92.6\% & 5.14 & 90.1\% & 5.89 & 68.2\%  \\
    \ding{51} & \ding{51} &  &2.85  &93.0\%  &5.28  &91.3\%  & 5.89 & 68.2\% \\
    \ding{51} &  & \ding{51} & 3.25 & 92.8\%  & 5.64 &91.4\%  & 6.01 & 90.4\% \\
    \ding{51} & \ding{51} & \ding{51} & 3.00 & 93.3\%  & 5.48 & 92.1\%  & 6.01 & 90.4\%  \\
    \bottomrule
    \end{tabular}
    }
    \label{tab:abla1}
\end{table*}

\section{Ablation Study}

\noindent \textbf{Ablation Study on Training and Inference Strategies.}
Table~\ref{tab:abla1} presents a comprehensive ablation study of both our training and inference designs. We compare different combinations of training and decoding strategies on GSM8K under three decoding thresholds, $\tau_{\mathrm{dec}} \in \{0.95, 0.5, 0.0\}$. On-policy rollout is the core of our training method. Even with OPUT alone, the model acquires the ability to revise its own errors, yielding substantial accuracy gains over the original model at $\tau_{\mathrm{dec}}=0.5$ and $0.0$. Our proposed SPD further improves robustness when many erroneous predictions emerge simultaneously, allowing the model to remain stable under highly parallel decoding and to preserve strong performance even in the extreme case of $\tau_{\mathrm{dec}}=0.0$. The key ingredient of SPD is to use soft embeddings, rather than discrete tokens, as intermediate decoding states. Maintaining the non-masked region as a contiguous prefix further improves performance. Another important result is that OPUT is a prerequisite for SPD. As shown in Table~\ref{tab:abla1}, directly applying SPD to the original model causes generation to collapse. This is because OPUT trains the model to recover clean tokens from both mask tokens and predicted tokens, making interpolation between their embeddings a meaningful and effective input for denoising.

\begin{wraptable}{r}{0.61\textwidth}
    \vspace{-0.8em}
    \centering
    \caption{Ablation study on block-level convergence criteria. The decoding threshold is set to 0.5.}
    \resizebox{0.61\textwidth}{!}{
    \begin{tabular}{c c c c c c}
    \toprule
    \multicolumn{2}{c}{\textbf{Convergence}} & 
    \multicolumn{2}{c}{\textbf{GSM8K}} & 
    \multicolumn{2}{c}{\textbf{MBPP}} \\
    \cmidrule(lr){1-2} \cmidrule(lr){3-4} \cmidrule(lr){5-6}
    \textbf{Consistency} &
    \textbf{Confidence} &
    \textbf{TPF $\uparrow$} &
    \textbf{Acc. $\uparrow$} &
    \textbf{TPF $\uparrow$} &
    \textbf{Acc. $\uparrow$} \\
    \midrule
    \ding{51} &  & 5.13 & 92.1\% & 5.16 & 79.9\% \\
     & \ding{51} & 2.28 & 92.2\% & 3.36 & 80.1\% \\
    \ding{51} & \ding{51} & 5.48 & 92.1\% & 5.86 & 79.2\% \\
    \bottomrule
    \end{tabular}
    }
    \label{tab:abla2}
    \vspace{-1.0em}
\end{wraptable}

\noindent \textbf{Ablation Study on Convergence Criteria.}
We further study in Table~\ref{tab:abla2} how different block-level convergence criteria affect the efficiency--performance trade-off. We consider two criteria: (1) \emph{consistency}, where decoding is considered converged if the model produces the same top-1 prediction for the block in two consecutive steps; and (2) \emph{confidence}, where decoding is considered converged if the confidence of every token in the block exceeds 0.9. As shown in Table~\ref{tab:abla2}, consistency serves as the primary convergence signal, with most blocks terminating once this condition is met. Adding the confidence criterion can further improve TPF by allowing decoding to stop before two consecutive identical predictions are observed, thereby saving the final forward pass. Importantly, neither criterion affects the accuracy.

\section{Related Work}

\noindent \textbf{Diffusion Language Models.}
Diffusion models \cite{ho2020denoising, song2020denoising} have become dominant in visual generation \cite{rombach2022high, podell2023sdxl, ruiz2023dreambooth, zhang2023adding}, and recent work has explored their application to text generation. Among existing paradigms, masked diffusion language models (MDLMs) \cite{shi2024simplified, austin2021structured,sahoo2024simple,zheng2024masked, lou2023discrete} have emerged as a promising alternative to AR-LLMs by modeling language in discrete space through masked token prediction. Building on this formulation, LLaDA \cite{nie2025large} and Dream \cite{ye2025dream} scale MDLMs to the billion-parameter regime with large-scale pretraining, demonstrating their practical potential. LLaDA-2.0 \cite{bie2025llada2} and LLaDA-MoE \cite{zhu2025llada2} further show that MDLMs can be effectively scaled with mixture-of-experts architectures. Beyond these developments, dLLMs are also attracting increasing attention in reasoning \cite{zhu2025llada,pan2025d,xie2025step,rojas2025improving,tang2025wd1,ni2026flexibility,zhao2025d1}, multimodal tasks \cite{you2025llada,yu2025dimple,yang2025mmada,ye2025dream2,liu2026mmada,wen2025llada,zeng2025diffusionvl,cheng2025sdar2}, code generation \cite{xie2025dream,gong2025diffucoder,fan2026stable}, long-context modeling \cite{liu2026longllada,he2025ultrallada,zheng2026mosaic}, and agent \cite{zhen2026dllm,zhao2026dllm}.

\noindent \textbf{Accelerating Diffusion Language Models.}
dLLMs are viewed as promising due to their potential for low-cost inference, yet their efficiency remains largely underexplored. Existing efforts improve efficiency from several perspectives. Some methods reduce the cost of each decoding step through techniques including KV caching \cite{ma2025dkv, liu2025dllm, wu2025fast,hu2025accelerating,liu2025wedlm}, token dropping \cite{chen2025dpad,huang2025mask,song2026sparse,xiao2026streaming}, and sparse attention \cite{wang2026sparsed,christoforos2025moe}. Others design more effective decoding strategies \cite{israel2025accelerating, wei2025accelerating, li2025beyond, gwak2025reward, ben2025accelerated,hong2025wide,hu2026residual,xu2025lopa, long2026focus,chen2026dflash, wang2025creditdecoding, qihierarchy,feng2026dvoting} to improve generation efficiency. A separate line of work \cite{song2025seed,qian2026d3llm,zhang2026t3d,bao2025learning,chen2025dultra,hu2026lightningrl} learns better decoding trajectories so that fewer decoding steps are required. dParallel \cite{chen2025dparallel} employs certainty-forcing distillation to accelerate confidence convergence and enable higher parallel decoding. Other methods \cite{wu2025fast2,cheng2025sdar,wang2025diffusion,arriola2025block,ma2026diffusion,tian2025next,liu2025tidar,fu2025efficient} interpolate between diffusion and autoregressive language models to better balance speed and accuracy. \cite{bie2026llada2,von2025generalized,zhang2025corrective} implement uniform training, which trains the model to recover clean tokens from random noisy tokens, thereby enabling token correction during generation. SM \cite{hersche2025soft} and EvoToken \cite{zhong2026beyond} introduce soft embeddings into the decoding process, but neither method translates this design into improved decoding efficiency. Further efforts \cite{xu2025dllmquant} leverage compression techniques to construct lightweight dLLMs.

\section{Conclusion}
In this paper, we present DMax, a novel paradigm for efficient diffusion language models that mitigates error accumulation for parallel decoding. DMax enables aggressive decoding parallelism while preserving the accuracy of the original model. We introduce two key components of our approach, namely On-Policy Uniform Training and Soft Parallel Decoding, and demonstrate their effectiveness through extensive experiments on diverse benchmarks. Our results establish a strong new baseline for parallel decoding in dLLMs and suggest a promising new direction for dLLMs.

\newpage
{
\small
\bibliographystyle{plain}
\bibliography{main}
}




\end{document}